\documentclass[a4paper]{article}

\usepackage[utf8]{inputenc}
\usepackage{erk}
\usepackage{times}
\usepackage{graphicx}
\usepackage[top=22.5mm, bottom=22.5mm, left=22.5mm, right=22.5mm]{geometry}

\usepackage[slovene,english]{babel}

\usepackage{bm}
\usepackage{amsmath}
\usepackage[breaklinks=true,letterpaper=true,colorlinks,bookmarks=false]{hyperref}
\usepackage{cleveref}

\usepackage{microtype}

\DeclareMathOperator*{\argmax}{arg\,max}
\DeclareMathOperator*{\argmin}{arg\,min}

\def\footnotemark{}

\begin{document}
\title{Hierarchical Superquadric Decomposition with \\Implicit Space Separation}

\author{Jaka Šircelj$^{1,2}$, Peter Peer$^{1}$, Franc Solina$^{1}$, Vitomir Štruc$^{2}$} 

\affiliation{$^{1}$Faculty of Computer and Information Science, UNI-LJ, Večna pot 113, Ljubljana, Slovenia\\
$^{2}$Faculty of Electrical Engineering, UNI-LJ, Tržaška cesta 25, Ljubljana, Slovenia\vspace{-3mm}
}\vspace{-3mm}


\maketitle

\begin{abstract}{Abstract}  
We introduce a new method to reconstruct 3D objects using a set of volumetric primitives, i.e., superquadrics. The method hierarchically decomposes a target 3D object into pairs of superquadrics recovering finer and finer details. While such hierarchical methods have been studied  before, we introduce a new way of splitting the object space using only properties of the predicted superquadrics. The method is trained and evaluated on the ShapeNet dataset. The results of our experiments suggest that reasonable reconstructions can be obtained with the proposed approach for a diverse set of objects with complex geometry.  

\end{abstract}

\selectlanguage{english}

\section{Introduction}

A vital field in 3D vision and robotics is 3D object reconstruction. By compressing input data into simpler representations such as meshgrids or sets of simpler objects, we can understand and interact with objects and their surroundings more easily, solving problems like collision avoidance~\cite{smith2020incorporating} or grasp planning~\cite{goldfeder2007grasp},~\cite{vezzani2017grasping}.

A common type of reconstruction is to represent a given (3D) object with a set of simpler shape primitives, often referred to as \textit{geons}. 
A popular approach in geon reconstruction is to estimate a fixed amount of geons and, concurrently, predict which of those should be kept in the final geon set, thus retaining the variability in geon numbers~\cite{tulsiani2017learning},~\cite{paschalidou2019superquadrics}. In our previous work, for example, we used a MaskRCNN model to first segment the input depth image and then reconstruct the parts using a specific type of geons, called superquadrics~\cite{sircelj2020segmentation}. 

\begin{figure}[t!]
    \begin{center}
        \includegraphics[width=0.85\linewidth]{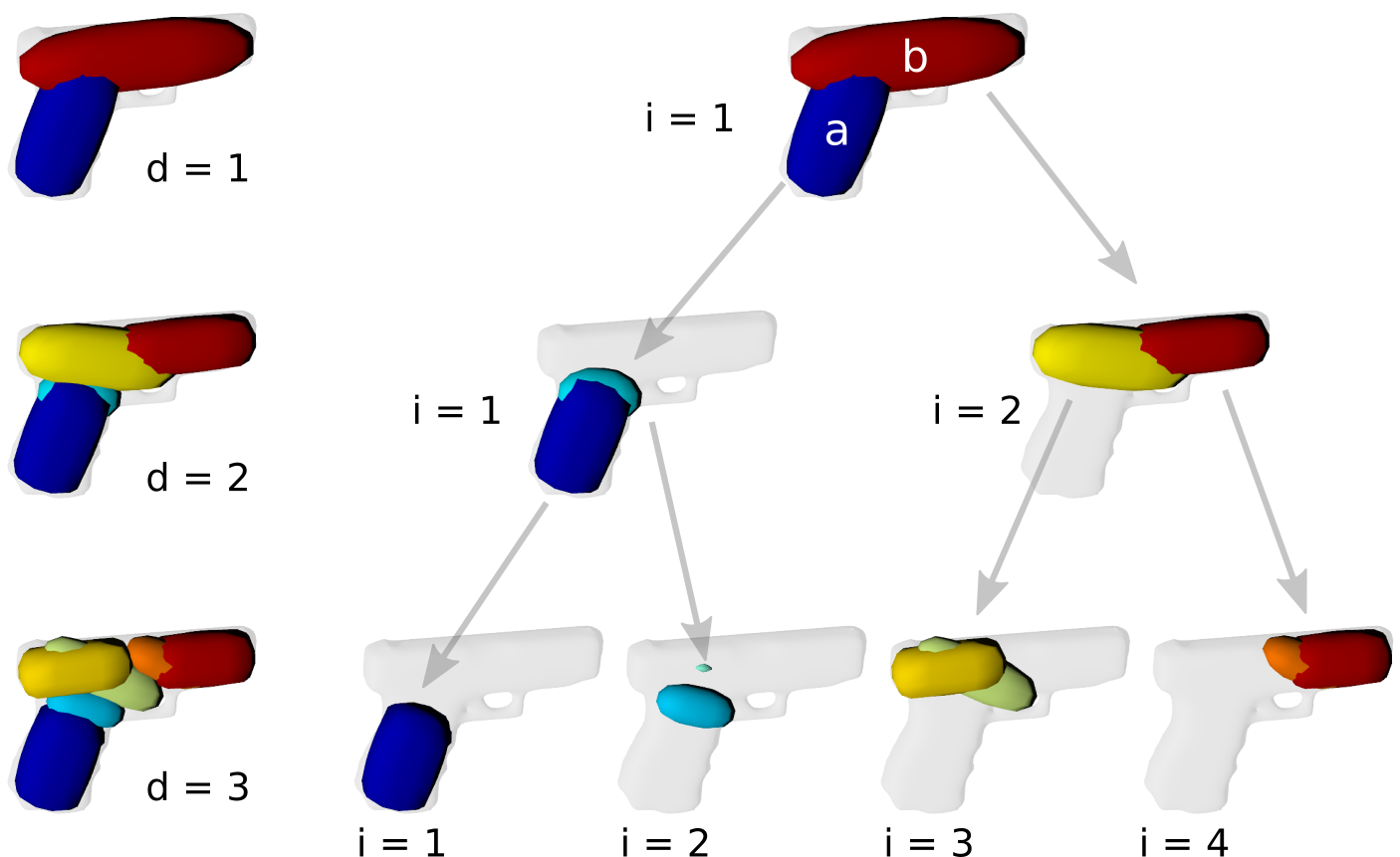}
        \caption{Visualization of the hierarchical decomposition with superquadrics. Each step splits the object into a pair of superquadrics ($a$ and $b$). The split is driven by previous superquadrics as shown by the arrows. The tree levels are indexed with $d$, whereas the SQ pairs in each level are indexed with $i$.\vspace{-3mm}}
        \label{fig:tree_examp}
    \end{center}
\end{figure}

Following the overall idea of our prior work, we introduce a novel method to reconstruct 3D objects with a set of superquadrics (SQ) in this paper. However, in this novel approach we base the procedure on a hierarchical tree decomposition algorithm. Instead of using a model that returns a fixed amount of geons at once, we introduce a hierarchical decomposition model, that incrementally splits the input object into more and more SQ representations, as illustrated in~\Cref{fig:tree_examp}. While a similar method was introduced by Paschalidou \textit{et al.} in~\cite{paschalidou2020learning}, we propose a splitting method based on the superquadrics characteristic, alleviating the model from determining how the object should be hierarchically split.

We evaluate the proposed approach on the ShapeNet dataset. Two models with different reconstruction capabilities are trained for the experiments. The first, facilitating reconstructions with a maximum of $4$ superquadrics, is trained and tested on the full ShapeNet dataset. The second, enabling reconstructions with up to $8$ superquadrics, is assessed on the pistol ShapeNet subset. Results are presented in terms of IoU scores and with visual examples, and point to the feasability of the proposed solution.

\section{Method}

\subsection{Superquadrics}

To reconstruct the 3D shapes we use superquadrics, which are geometric shapes that can describe objects such as spheres, ellipsoids, cylinders and rectangular cuboids. A common way to describe the surface of a superquadric is using the implicit function $F(x, y, z) = 1$. Here $F$ is called the \textit{inside-outside function} and is defined as 
\begin{equation} \label{eq:sq_function}
    F(x, y, z) = \left( 
        \left( \frac{x}{a_1} \right)^\frac{2}{\varepsilon_2} + 
        \left( \frac{y}{a_2} \right)^\frac{2}{\varepsilon_2} 
    \right)^\frac{\varepsilon_2}{\varepsilon_1} + 
    \left( \frac{z}{a_3} \right)^\frac{2}{\varepsilon_1},
\end{equation}
where $a_1, a_2, a_3$ define the size of the superquadric and $\varepsilon_1, \varepsilon_2$ its shape. We can also move the superquadric in space by offsetting the $x, y, z$ coordinates by $t_1, t_2, t_3$ respectively and by rotating the coordinates using quaternion notation $q_0, q_1, q_2, q_3$. To abbreviate the notation we write the spatial coordinates using vector notation $\bm{x} = [x, y, z]$, and all superquadric parameter as 
$$\bm\lambda = [a_1, a_2, a_3, \varepsilon_1, \varepsilon_2, t_1, t_2, t_3, q_1, q_2, q_3, q_4].$$

A useful property of the \textit{inside-outside function}, that will be used further on, is that it contains information whether the space is inside or outside the superquadric (hence the name). Points with $F(\bm{x}) < 1$ lie inside the object, $F(\bm{x}) > 1$ lie outside, and, as already suggested, points with $F(\bm{x}) = 1$ lie on its surface. For numerical stability we rather compare $F(\bm{x})^{\varepsilon_1}$ values~\cite{jaklic2000segmentation}, the spatial inequalities remains the same.
\begin{figure}[!t]
    \begin{center}
        \includegraphics[width=0.88\linewidth]{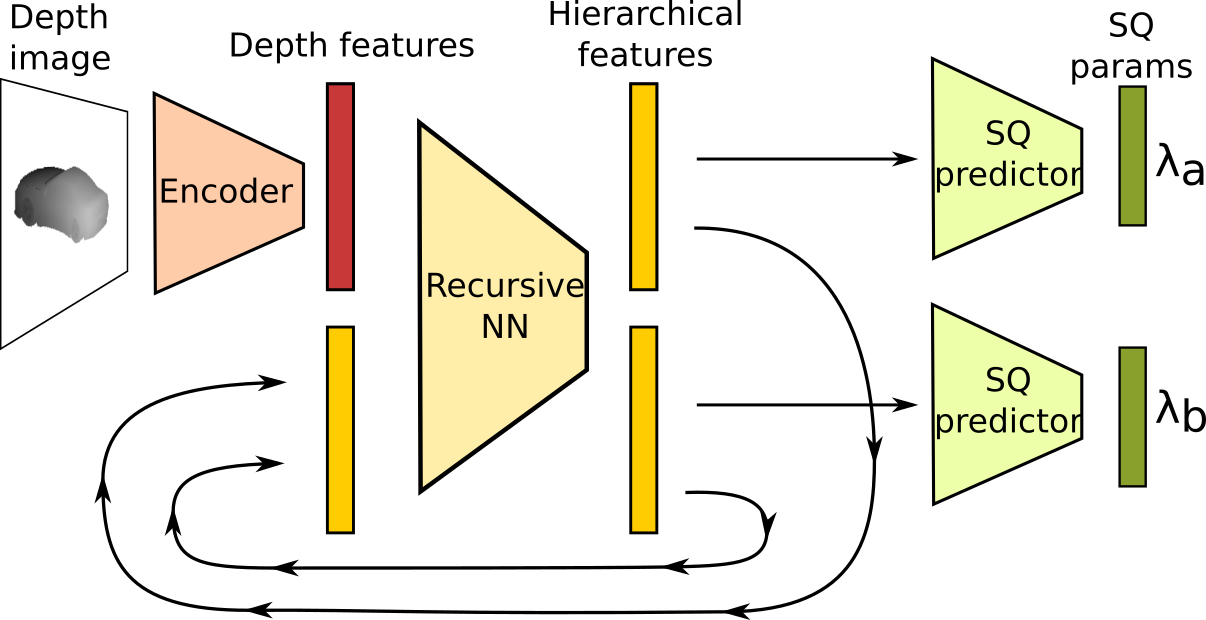}
        \caption{Recursive model architecture.}\vspace{-4mm}
        \label{fig:model_arch}
    \end{center}
\end{figure}

\subsection{Model architecture}

Similarly to Paschalidou \textit{et al.}~\cite{paschalidou2020learning} we use a hierarchical decomposition procedure to reconstruct a complex 3D object using a set of superquadrics. This allows the decomposition to predict different numbers of superquadrics, increasing their number for fine detail, and decreasing for larger parts of the object. The procedure works by first extracting a feature vector out of the input depth image, which we refer to as a \textit{depth feature}. This is concatenated with a \textit{hierarchical feature} and passed into a recursive neural network which predicts two new \textit{hierarchical features} each encoding data to predict a new superquadric. This prediction is done with an \textit{SQ predictor} network which predicts the parameters $\bm{\lambda}$ previously described. Each of the two \textit{hierarchical features} can again be concatenated with the depth feature vector and passed into the recursive neural network, thus splitting the initial superquadric feature into two. A hierarchical feature in the first step can be a simple vector filled with zeroes. The model architecture corresponding to the outlined idea is shown in \Cref{fig:model_arch}.

Such a model produces a superquadric-pair binary tree, where each node contains two sets of parameters and two links to two children, again containing two sets of parameters. We use the following notation to reference the superquadric parameters in the tree:
\begin{equation}
    \textrm{SQ\_pair}_{d, i} = [\bm{\lambda}_a, \bm{\lambda}_b],
    \label{eq:sq-pair-tree}
\end{equation}
where $d$ is the depth of the node and $i$ is the index of the node in the tree. This notation is also shown in the tree example in \Cref{fig:tree_examp}.


\subsection{Training}

For training, we use the natural property of the \textit{inside-outside} function, whose sign describes the inside and the outside space of the superquadric. Similarly to \cite{paschalidou2020learning}, we define the occupancy function $g(\bm x; \bm\lambda)$, which translates this property to act as a binary classifier of the inside space
\begin{equation}
    g(\bm x; \bm\lambda) = \sigma(s * (1 - F^{\,\epsilon_1}(\bm{x}; \bm\lambda))),
\end{equation}
where $\sigma$ is the sigmoid function, and $s$ is a scaling parameter that defines the slant of the values around the surface of the superquadric -- for more information see~\cite{paschalidou2020learning}.

Since we want to use this occupancy property of the superquadrics, we generate the training dataset by sampling points in space and annotating them with $1$ if they lie inside the 3D object and $0$ if they lie outside. The root node of the superquadric pair tree, Eq.~\eqref{eq:sq-pair-tree}, is trained on these sets of initial ground-truth points and labels, denoted as $\mathcal{P}$ and $\mathcal{L}_{1,1}$ respectively.

\begin{figure}[!t]
    \begin{center}
        \includegraphics[width=\linewidth]{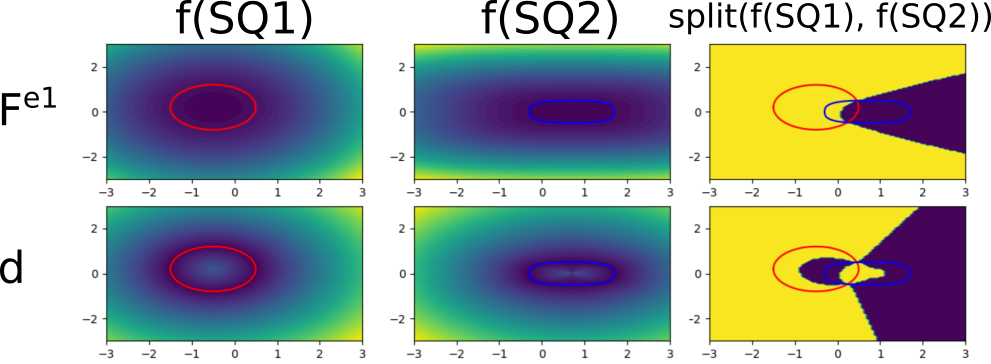}
        \caption{Rows: top - computations on the $f=F^{\varepsilon_1}$ function, bottom - computations on the radial Euclidean distance function ($f=d$). Columns: first - values of the $f$ function around the first superquadric, shown in red, second - values of $f$ around the second superquadric, shown in blue, third - space split according to the $f$ function. We can see that $F$ and $F^{\varepsilon_1}$ split space better inside of the superquadrics while the radial distance function splits space more evenly outside of the superquadrics.}\vspace{-5mm}
        \label{fig:sq_split}
    \end{center}
\end{figure}

The child SQ pairs should only cover the space in proximity to the space occupied by their parent superquadric, and should not be fitted to the full inside of the 3D model i.e.\ $\mathcal{L}_{1,1}$. Paschalidou \textit{et al.}~\cite{paschalidou2020learning}  solve this problem by also predicting the centroid of the part covered by a predicted superquadric. The space is then split by matching every point to its closest centroid (see Figure 4 in~\cite{paschalidou2020learning}).
In this paper, we use the mathematical properties of superquadrics to split the space. More specifically, we found that it is best to split the space inside and outside of the superquadric pairs differently. The space inside of the superquadrics is split using the \textit{inside-outside} function by finding the maximum of the \textit{inside-outside} function, i.e.:
\begin{equation}
    \argmax_{i = a,b} F^{\varepsilon_1}(\bm x; \bm\lambda_i).
\end{equation}
Conversely, the space outside of the superquadric pair is split using the radial Euclidean distance
\begin{equation}
    d(\bm x)= |\bm x|| 1 - F^{-\frac{\varepsilon_1}{2}} |,
\end{equation}
taking the superquadric with the minimum value
\begin{equation}
\argmin_{i = a,b} d_i(\bm x).    
\end{equation}
This distance function is used outside because it behaves more properly as a distance function further away from a superquadric. For more information on the radial distance function, look at Jaklič \textit{et al.}~\cite{jaklic2000segmentation}. The space-splitting properties of the \textit{inside-outside} function and the \textit{radial Euclidean} distance function are illustrated in \Cref{fig:sq_split}. The proposed splitting method is simpler than the one used in~\cite{paschalidou2020learning}, since we don't need to predict any additional values, and uses the space splitting capabilities of the superquadrics, thus, splitting the space more naturally and in accordance with the geometry of the superquadrics.


\begin{figure}[!t]
    \begin{center}
        \includegraphics[width=0.9\linewidth]{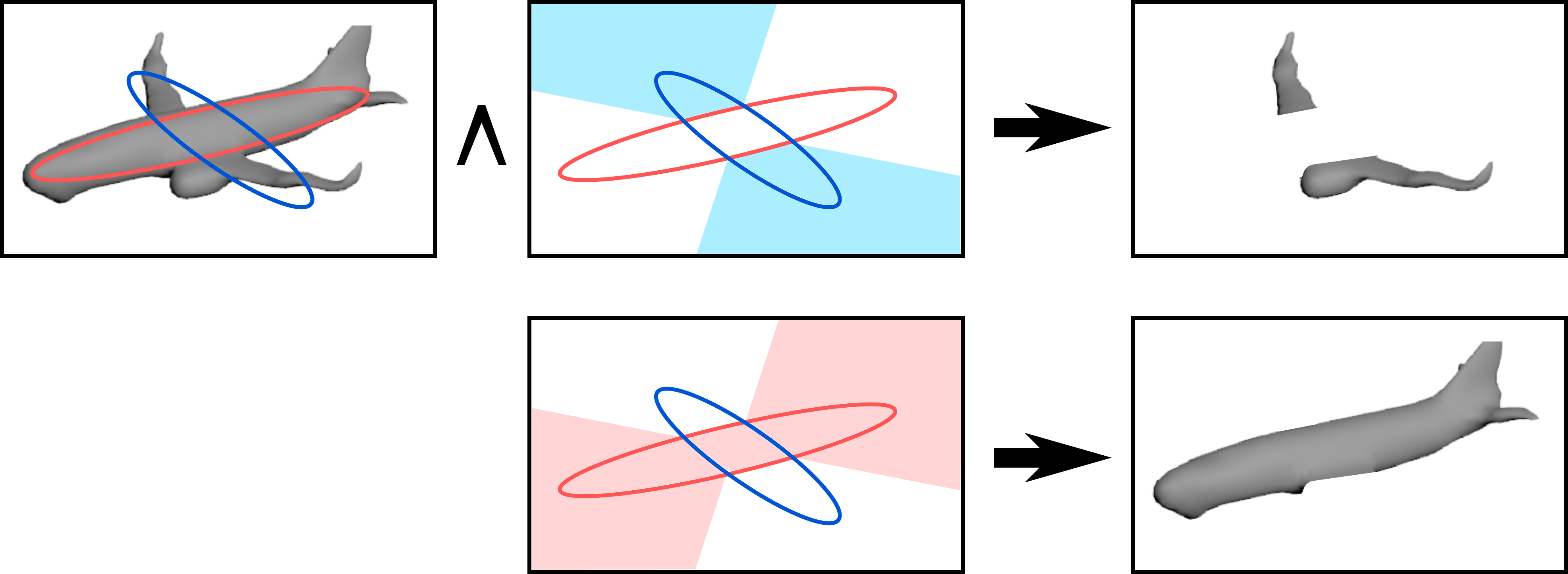}
        \caption{Left: Airplane with its insides noted dark $\mathcal{L}_{1,1}$, center top: outline of SQ 1 with its assigned space after split, center bottom: outline of SQ 2 with its assigned space, right: two outcomes of \Cref{eq:space_spilt}, top is $\mathcal{L}_{2,1}$, bottom $\mathcal{L}_{2,2}$.}
        \label{fig:obj_inside_outside}
    \end{center}\vspace{-4mm}
\end{figure}

Since our model obtains a hierarchy of superquadrics, we must also construct a hierarchy of labels $\mathcal{L}_{d,i}$ on which we compute the losses. These labels are computed recursively according to how the space is split using the superquadric pairs. The inside of the object is considered inside for the SQ-pair child only if its parent superquadrics' split space also contains it (see \Cref{fig:obj_inside_outside}). This translates to a logical AND operation between the parents ground truth occupancy $\mathcal{L}_{d,i}$ and the space split of the parent 
\begin{equation}
\begin{split}
    l_{d, i} =& \;  l_{parent\_node(d, i)} \;\; \wedge \\
    & \; split(\bm x, \bm\lambda_{parent\_sq(d, i)}, \bm\lambda_{uncle\_sq(d, i)}),
\end{split}
\label{eq:space_spilt}
\end{equation}
where $l_{d, i} \in \mathcal{L}_{d, i}$ is the label belonging to point $\bm x \in \mathcal{X}$, $parent\_node(d, i)$ denotes the parent SQ pair node, $parent\_sq(d, i)$ is the parent superquadric ($a$ or $b$) and $uncle\_sq(d, i)$ is the other superquadric from the parent pair node. 
For example, for node $(d, i) = (2, 2)$:
\begin{equation*}
\begin{split}
    parent\_node(2, 2) &= (1, 1)\\
    parent\_sq(2, 2) &= (1, 1, a)\\
    uncle\_sq(2, 2) &= (1, 1, b).
\end{split}    
\end{equation*}

\begin{table}[!t]
\caption{IoU scores of ModelS (full ShapeNet).} \label{tab:models_iou}
\smallskip
\small
\begin{center}
\begin{tabular}{ | r | c | c | }
\hline  
  \textbf{SQ tree level} & $1$ & $2$\\
  \hline 
  \textbf{IoU} [in \%] & $56.7\%$ & $58.8\%$ \\
\hline
\end{tabular}\vspace{-5mm}
\end{center}
\end{table}
\begin{table}[t]
\caption{Per object category IoU scores of ModelS (full ShapeNet). Superquadrics taken from the last/second tree layer.} \label{tab:iou_shapenet_sep}
\begin{center}
\small
\begin{tabular}{ | r | c || r | c | }
\hline
 \textbf{Label}       &        \textbf{IoU} [in \%] &       \textbf{Label}  & \textbf{IoU} [in \%]        \\ \hline\hline
 dishwasher  & $ 86.9 \%$ &      rocket  & $ 64.7 \%$ \\ \hline
  microwave  & $ 84.3 \%$ &     printer  & $ 64.2 \%$ \\ \hline
        bus  & $ 82.1 \%$ &     monitor  & $ 62.9 \%$ \\ \hline
     washer  & $ 81.2 \%$ &  watercraft  & $ 61.9 \%$ \\ \hline
       file  & $ 80.4 \%$ &         bag  & $ 61.4 \%$ \\ \hline
        can  & $ 78.1 \%$ &       piano  & $ 60.4 \%$ \\ \hline
     pillow  & $ 77.9 \%$ &       plane  & $ 57.6 \%$ \\ \hline
        car  & $ 76.8 \%$ &       rifle  & $ 52.2 \%$ \\ \hline
      phone  & $ 76.1 \%$ &       knife  & $ 50.8 \%$ \\ \hline
     laptop  & $ 73.1 \%$ &       chair  & $ 47.5 \%$ \\ \hline
    cabinet  & $ 71.8 \%$ &       table  & $ 47.4 \%$ \\ \hline
    speaker  & $ 69.9 \%$ &       bench  & $ 46.7 \%$ \\ \hline
       sofa  & $ 68.8 \%$ &        lamp  & $ 41.9 \%$ \\ \hline
     camera  & $ 67.4 \%$ &     bathtub  & $ 41.2 \%$ \\ \hline
     pistol  & $ 67.2 \%$ &    bookcase  & $ 40.2 \%$ \\ \hline
    mailbox  & $ 65.3 \%$ &        bowl  & $ 34.3 \%$ \\ \hline
     basket  & $ 23.8 \%$ &              &            \\ \hline
\end{tabular}
\end{center}\vspace{-6mm}
\end{table}
\begin{table}[h]
\caption{IoU scores of ModelP (pistols).} \label{tab:modelp_iou}
\smallskip
\small
\begin{center}
\begin{tabular}{ | r | c | c | c |}
\hline  
  \textbf{SQ tree level} & $1$ & $2$ & $3$ \\
  \hline 
  \textbf{IoU} & $63.5\%$ & $65.6\%$ & $64.7\%$ \\
\hline
\end{tabular}\vspace{-4mm}
\end{center}
\end{table}

Finally the occupancy loss is calculated using the occupancy values for each pair tree node and the ground truth occupancy values $\mathcal{L}_{d,i}$
\begin{equation}
    L = \sum_{d=1} \sum_{i=1}^{2^{d-1}}  \sum_{\substack{\bm x \in \mathcal{X}\\l_{d,i} \in \mathcal{L}_{d,i}}} L_{BCE}(\max_{sq = {a,b}} g(\bm x; \bm\lambda_{sq}), l_{d, i}),
\end{equation}
where $L_{BCE}$ is the binary cross entropy loss. This loss is similar to the \textit{part-reconstruction loss} term from Paschalidou \textit{et al.}~\cite{paschalidou2020learning}, but differs in that we look at how accurately an SQ pair matches the designated space of its parent superquadric.




\section{Results}

In this section, we present the results for two trained hierarchic decomposition models. We use the ShapeNet dataset~\cite{shapenet2015} for training along with the NVIDIA Kaolin v0.1 library~\cite{KaolinLibrary}, which is used to compute signed distance function values of the 3D objects and to infer the ground truth inside-outside space. The first model (abbreviated ModelP) is trained on the Pistol subset of ShapeNet with the maximum depth of the superquadric pair tree set to $3$ levels. The second model (ModelS) is trained on the full ShapeNet dataset with the maximum depth set to $2$ levels. Training with more levels proved challenging, and resulted in degenerated SQs, often small and out of bounds.

\textbf{Performance Indicator.} After a train-validation-test split, both models are evaluated on the test subset using the intersection-over-union metric (IoU)
\begin{equation}
\label{eq:iou}
    IoU(\Lambda_d, \mathcal{L}_{0,0}) = 
    \frac{\sum_{\substack{\bm x \in \mathcal{X}\\l \in \mathcal{L}_{0,0}}} \hat l_{d}(\bm x) \wedge l}
    {\sum_{\substack{\bm x \in \mathcal{X}\\l \in \mathcal{L}_{0,0}}} \hat l_{d}(\bm x) \vee l },
\end{equation}
where $\Lambda_d$ denotes all predicted superquadric parameters in the tree at layer $d$, $\Lambda_d = \{\lambda_{d,i} | i \in \left[ 1, 2^{d-1} \right]\}$ and $\hat l_{d}(\bm x)$ denotes the predicted inside-outside label for point $\bm x$. The latter is inferred as inside or $\hat l_{d}(\bm x)  = 1$ if $g(\bm x; \bm\lambda_{d,i}) > 0.5$, for any superquadric in layer $d$, otherwise the point is outside of the reconstruction.


\textbf{Model Comparison.} The IoU results of the ModelS on the full ShapeNet dataset are shown in~\Cref{tab:models_iou} and for separate subsets of Shapenet in~\Cref{tab:iou_shapenet_sep}. The IoU results of ModelP for the pistol subset are reported in~\Cref{tab:modelp_iou}. Since the full ShapeNet dataset contains much more variability, ModelS scores a lower IoU over the full ShapeNet dataset, than ModelP over the pistol subset. An interesting observation is that ModelS still performs better on the pistol subset than ModelP, even though the latter was trained exclusively on pistols. This suggests that training on more complex and diverse sets can improve results on specific subsets. However, training on the full ShapeNet did prove more difficult. For instance, we managed to train ModelS for only up to $2$ tree levels, while we managed to train ModelP to predict trees up to $3$ levels deep.

\textbf{Object-specific Results.} Regarding the specific subsets of ShapeNet, the model performed best on simpler object classes, e.g., \textit{dishwasher} or \textit{microwave}, which have a box-like shape and are, thus, easier to reconstruct using superquadrics. The poorest results are on concave objects, such as \textit{bowl} or \textit{bathtub}, since only two levels of SQ-pairs don't give enough capacity to cover the object.

\begin{figure}[!t]
    \begin{center}
        \includegraphics[width=0.8\linewidth]{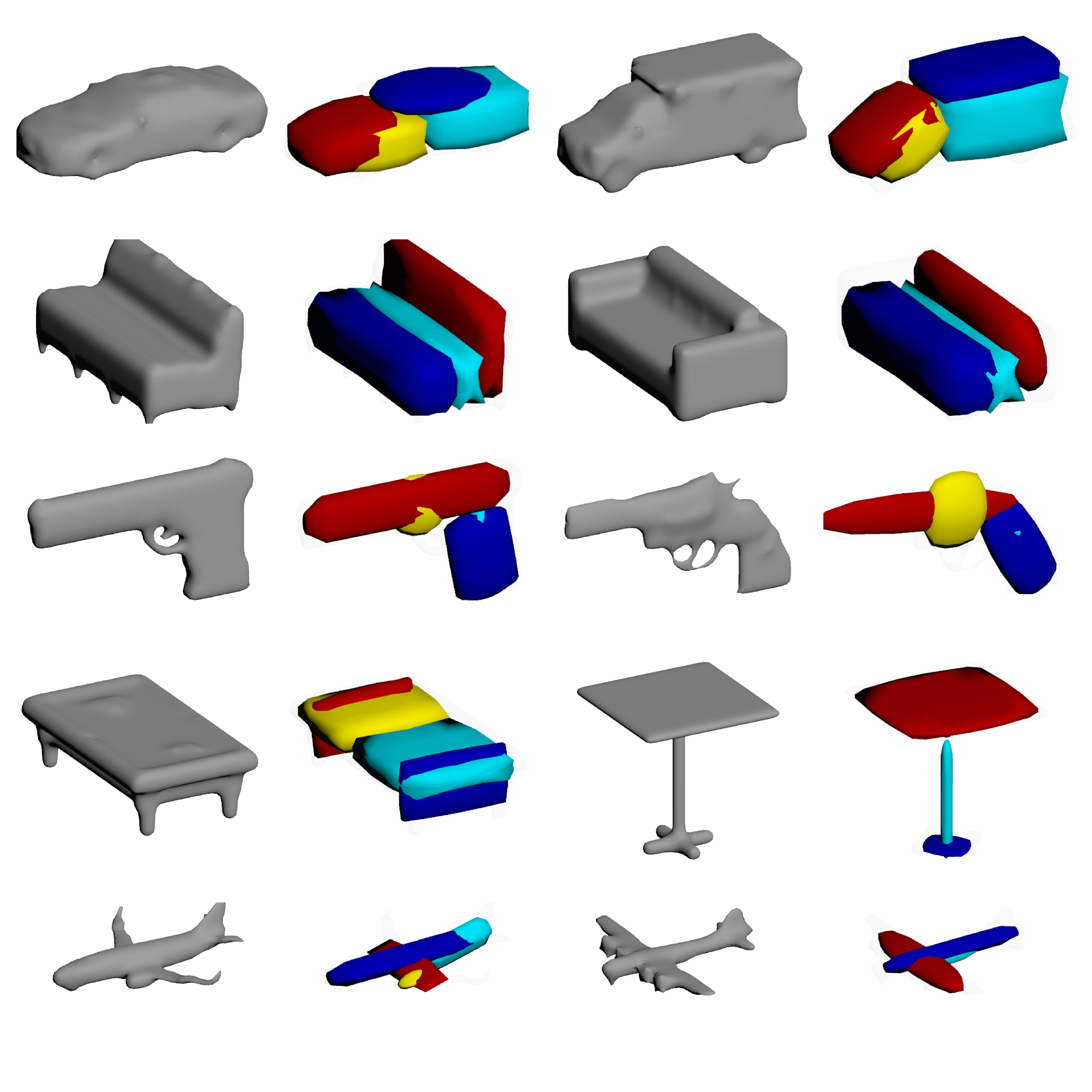} \vspace{-6mm}
        \caption{Reconstruction examples of ModelS on full ShapeNet. The 1st and 3rd columns show the original object. The 2nd and 4th show the reconstruction from the second SQ  tree layer.}
        \label{fig:shapenet_rec_vis}
    \end{center}\vspace{-6mm}
\end{figure}

\textbf{Qualitative Evaluation.} In~\Cref{fig:shapenet_rec_vis} we show a few reconstruction examples of ModelS. As we can see, the largest issue is the low number of predicted superquadrics, given that ModelS has been trained for only $2$ levels, and being capable to reproduce an object with a maximum of 4 superquadrics. While simpler objects like the two cars are well reconstructed, more complex objects prove impossible to reconstruct with $4$ superquadrics. 
For example, both tables have four leg stands, which get covered by two rectangular superquadrics in the left case, and a cylindrical one on the right. The same problem can be observed on both of the planes, where the wings and the engine prove much too complex for only 4 superquadrics.

\begin{figure}[!htb]
    \begin{center}
        \includegraphics[width=0.85\linewidth]{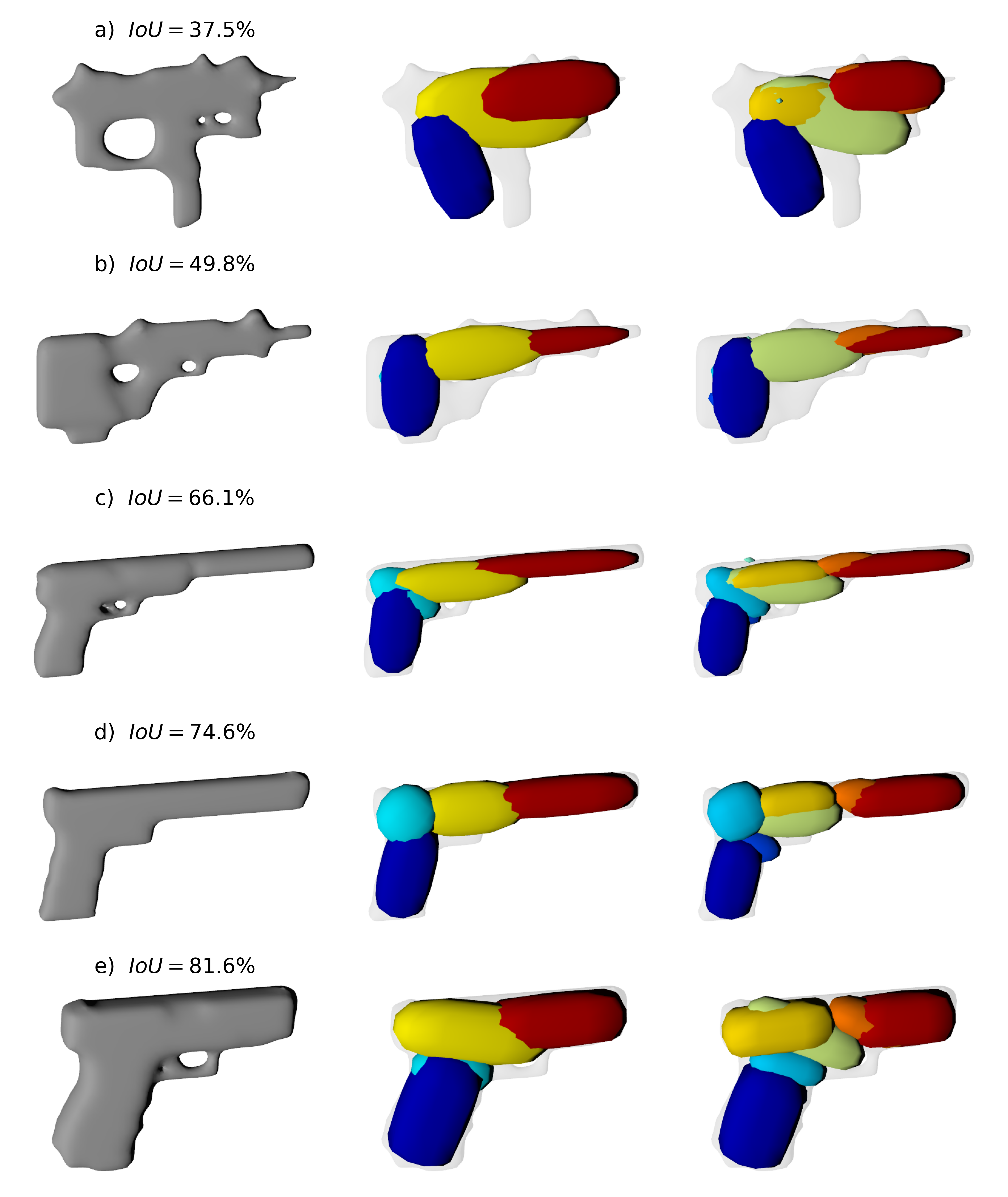}\vspace{-3mm}
        \caption{Reconstruction examples of ModelP on the pistol subset. First column - original object, second - reconstruction from the second SQ tree level, third - reconstruction from the third SQ tree level. The reconstructions are sorted from worst to best with the reconstruction IoU annotated above.}
        \label{fig:pistols_rec_vis}
    \end{center}\vspace{-6mm}
\end{figure}

In~\Cref{fig:pistols_rec_vis} we show a few ModelP reconstructions, from worst to best, using the $2^{nd}$ and $3^{rd}$ tree level. Here, we observe that the least performing objects were those pistol models that have been rare in the pistol subset, like the $1^{st}$st and $2^{nd}$ pistol. The most common handgun in the last two rows performed best in terms of IoU.

\section{Conclusion}

The hierarchical decomposition has proven to successfully decompose the targeted input objects, even with single point-of-view depth images, which give a highly occluded representation of the 3D object. We showed that, unlike in Paschalidou \textit{et al.}~\cite{paschalidou2020learning}, we don't need to predict centroids to split the space into two. It is sufficient to use the already predicted superquadrics for space splitting. These results have been supported by fairly good IoU scores and good visual representations. There is also much to improve in the model. The largest issue we faced was the maximum allowed depth of the SQ pair tree, where we achieved the max depth of $3$ with $8$ predicted superquadrics. 

\footnotesize
\bibliographystyle{ieee}
\bibliography{erk}

\end{document}